\documentclass[10pt]{extarticle}
\usepackage[backend=biber,sorting=ynt]{biblatex}
\usepackage{comment,url,algorithm,algorithmic,graphicx,subcaption,relsize}
\usepackage{amssymb,amsfonts,amsmath,amsthm,amscd,dsfont,mathrsfs,mathtools,nicefrac}
\usepackage{float,psfrag,epsfig,color,xcolor,url,hyperref}
\usepackage{epstopdf,bbm,mathtools,enumitem}

\def\balign#1\ealign{\begin{align}#1\end{align}}
\def\baligns#1\ealigns{\begin{align*}#1\end{align*}}
\def\balignat#1\ealign{\begin{alignat}#1\end{alignat}}
\def\balignats#1\ealigns{\begin{alignat*}#1\end{alignat*}}
\def\bitemize#1\eitemize{\begin{itemize}#1\end{itemize}}
\def\benumerate#1\eenumerate{\begin{enumerate}#1\end{enumerate}}

\newenvironment{talign*}
 {\csname align*\endcsname}
 {\endalign}
\newenvironment{talign}
 {\csname align\endcsname}
 {\endalign}

\def\balignst#1\ealignst{\begin{talign*}#1\end{talign*}}
\def\balignt#1\ealignt{\begin{talign}#1\end{talign}}

\let\originalleft\left
\let\originalright\right
\renewcommand{\left}{\mathopen{}\mathclose\bgroup\originalleft}
\renewcommand{\right}{\aftergroup\egroup\originalright}

\def\tinycitep*#1{{\tiny\citep*{#1}}}
\def\tinycitealt*#1{{\tiny\citealt*{#1}}}
\def\tinycite*#1{{\tiny\cite*{#1}}}
\def\smallcitep*#1{{\scriptsize\citep*{#1}}}
\def\smallcitealt*#1{{\scriptsize\citealt*{#1}}}
\def\smallcite*#1{{\scriptsize\cite*{#1}}}

\def\<{\left\langle} 
\def\>{\right\rangle}

\def\norm#1{\left\|{#1}\right\|} 
\newcommand{\twonorm}[1]{\norm{#1}_2} 
\ifdefined\nonewproofenvironments\else
\ifdefined\ispres\else

\newenvironment{proof-sketch}{\noindent\textbf{Proof Sketch}
  \hspace*{1em}}{\qed\bigskip\\}
\newenvironment{proof-idea}{\noindent\textbf{Proof Idea}
  \hspace*{1em}}{\qed\bigskip\\}
\newenvironment{proof-of-lemma}[1][{}]{\noindent\textbf{Proof of Lemma {#1}}
  \hspace*{1em}}{\qed\\}
\newenvironment{proof-of-theorem}[1][{}]{\noindent\textbf{Proof of Theorem {#1}}
  \hspace*{1em}}{\qed\\}
\newenvironment{proof-attempt}{\noindent\textbf{Proof Attempt}
  \hspace*{1em}}{\qed\bigskip\\}

\fi

\fi
\makeatletter
\@addtoreset{equation}{section}
\makeatother

\hypersetup{
  colorlinks,
  linkcolor={red!50!black},
  citecolor={blue!50!black},
  urlcolor={blue!80!black}
}

\newcommand{\handout}[5]{
  \noindent
  \begin{center}
    \framebox{
      \vbox{
        \hbox to 5.78in { {\bf \title } \hfill #2 }
        \vspace{4mm}
        \hbox to 5.78in { {\Large \hfill #5  \hfill} }
        \vspace{2mm}
        \hbox to 5.78in { {\em #3 \hfill #4} }
      }
    }
  \end{center}
  \vspace*{4mm}
}

\usepackage{setspace}
\usepackage{dirtytalk}

\usepackage{soul}

\usepackage{titlesec}

\titleformat{\section}{\normalfont\fontsize{10}{10}\bfseries}{\thesection}{1em}{}
\titleformat{\subsection}{\normalfont\fontsize{10}{10}\bfseries}{\thesubsection}{1em}{}

\usepackage{comment}

\usepackage{geometry}
\begin{document}


\begin{center}
	\textbf{{Neuro-Symbolic VQA: A review from the perspective of AGI desiderata}}\\
	Ian Berlot-Attwell
\end{center}

An ultimate goal of the AI and ML fields is artificial general intelligence (AGI); although such systems remain science fiction, various models exhibit aspects of AGI. In this work, we look at neuro-symbolic (NS) approaches to visual question answering (VQA) from the perspective of AGI desiderata. We see how well these systems meet these desiderata, and how the desiderata often pull the 
scientist in opposing directions. It is my hope that through this work we can temper model evaluation on benchmarks with a discussion of the properties of these systems and their potential for future extension.

\section{Introduction}

 VQA is the task of answering a natural language question about a given image. NS approaches combine neural networks with a symbolic component (e.g., symbolic reasoning, symbolic representations, or exploiting pre-existing symbolic knowledge). NS approaches to VQA are of particular interest as they may 
 promote
 compositional generalization \cite{probab_ns_vqa, 
 nmn, 
 ns_survey} 
 (i.e., the ability to understand new combinations of previously acquired concepts; for instance, 
 \say{Buzz Aldrin wielded a glass hammer}). 
 NS approaches also promise to be more interpretable \cite{TbD, nsvqa}, 
and may allow for simpler skill transfer between tasks.

Of the AGI desiderata identified by Bieger et al. \cite{agi}, we discuss: \textit{natural growth} (the ability to continue learning without human re-design), \textit{traceability} (the ability to give step-by-step explanations of decisions, even if step execution cannot be explained), \textit{transfer-learning} (the ability to exploit previous knowledge when learning new knowledge), \textit{few-shot learning}, and \textit{self-awareness of limitations}. For \textit{few-shot learning}, we mostly consider \textit{compositional generalization} (the ability to understand previously learned concepts in new combinations).  These desiderata were chosen as they can be found in neuro-symbolic VQA models. As these desiderata are meaningless if we do not take practical considerations into account, we also discuss \textit{performance}, \textit{scalability}, and \textit{training data} when they become particularly relevant.

In this work: \textit{Attribute} refers to properties of an object (e.g., colour, or shape), \textit{Concept} refers to possible values of an \textit{Attribute} (e.g., red, or circle), and \textit{Relation}s are possible relations between two or more objects (e.g., besides, under, or is wearing). A \textit{Domain Specific Language} (DSL) is a symbolic language that can be used to represent any question in the domain. For instance, the DSL \{\texttt{sphere}, \texttt{cube}, \texttt{left\_of($\cdot,\cdot$)}\} can represent \textit{Is there a cube left of a sphere?} as \texttt{left\_of(cube, sphere)}. We call the representation of a question in a given DSL, the \textit{Program}.

\begin{figure}[h!]
	\centering
	\includegraphics[width=0.5\linewidth]{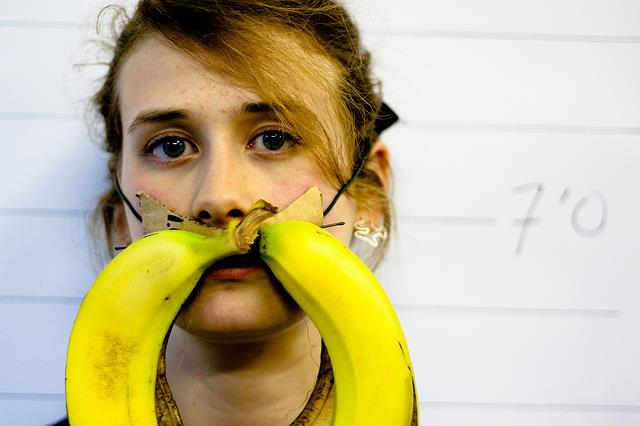}
	\caption{A sample image from the VQA 1.0 dataset \cite{VQA}; the corresponding questions include ``What color are her eyes?'' and ``What is the mustache made of?'' }
	\label{fig:vqa}
\end{figure}

\begin{figure}[h!]
	\centering
	\includegraphics[width=0.6\linewidth]{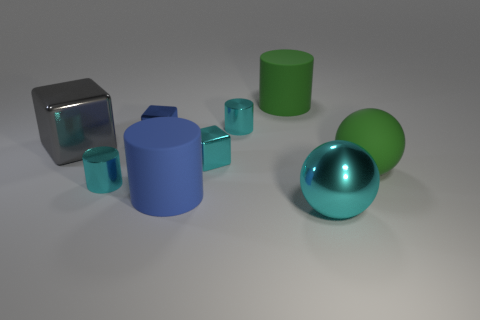}
	\caption{A sample image from the CLEVR dataset \cite{nmn}; the corresponding questions include ``Are there fewer metallic objects that 	are on the left side of the large cube than cylinders to the left of the cyan shiny block?'' and ``There is a green 	rubber thing that is left of the rubber thing that is right of the rubber cylinder behind the gray shiny block; what is its size?''}
	\label{fig:clevr}
\end{figure}

\begin{figure}[h!]
	\centering
	\includegraphics[width=0.5\linewidth]{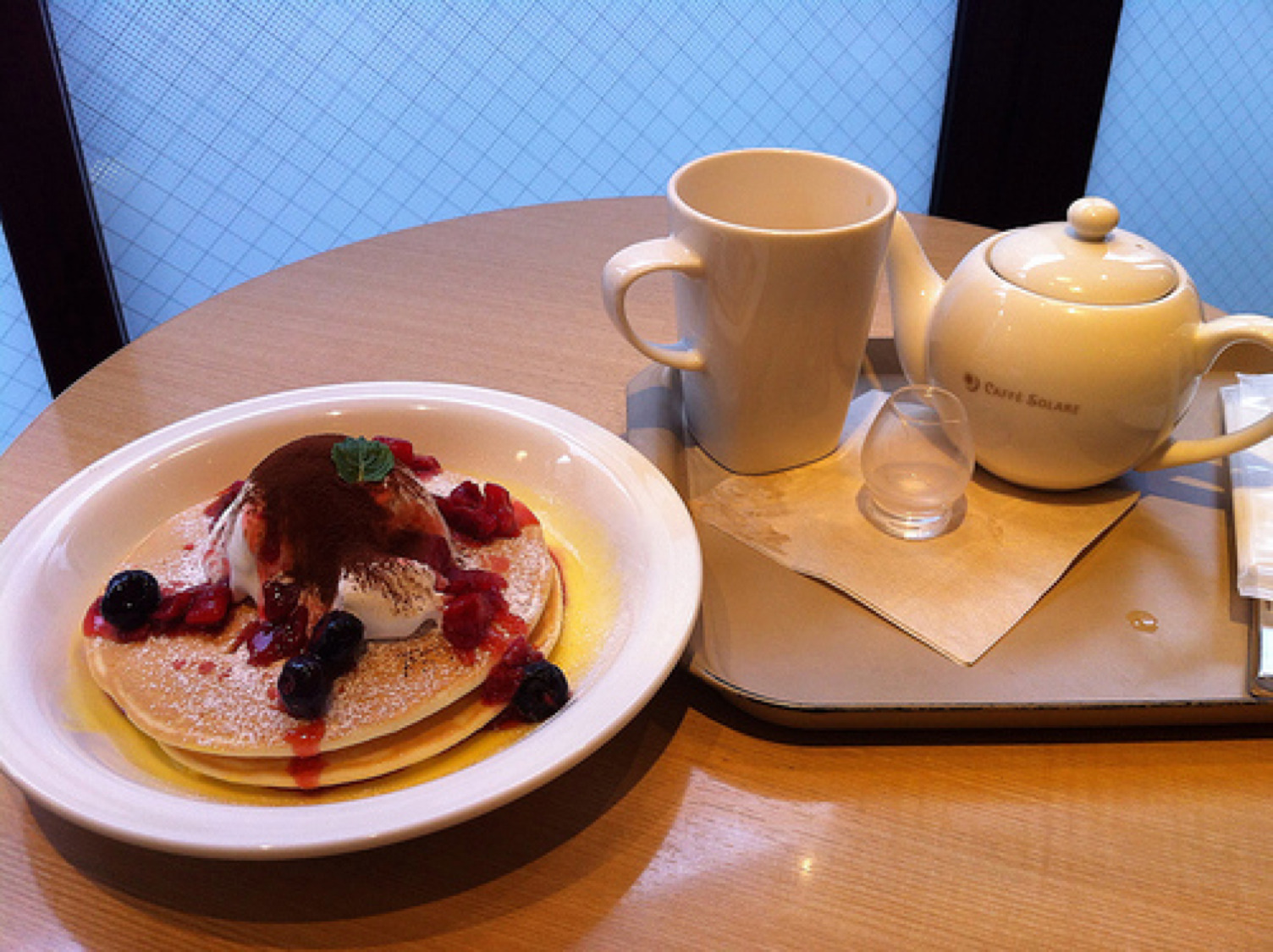}
	\caption{A sample image from the GQA dataset \cite{gqa}; the corresponding questions include ``Is there any fruit to the left of the tray the cup is on top of?'' and ``Are there any cups to the left of the tray on top of the table?''}
	\label{fig:gqa}
\end{figure}

Some frequently used datasets are: the VQA 1.0 dataset \cite{VQA} of natural images with human-created questions (see Fig \ref{fig:vqa}), the synthetic SHAPES dataset \cite{nmn} of shapes on a $3\times3$ grid with compositional questions (see Fig \ref{fig:nmnexec}, \ref{fig:shapes}), the CLEVR dataset \cite{clevr} which is similar to SHAPES but using 3D figures (see Fig \ref{fig:clevr}), and the GQA dataset \cite{gqa} of natural images  
with synthetic compositional questions (see Fig \ref{fig:gqa}). A variant of CLEVR, CLEVR-CoGenT \cite{clevr}, is also used to test \textit{compositional generalization}; in this dataset, cubes and cylinders are different colours and swap colour palettes between train and test time.

\setlength{\abovecaptionskip}{0pt}
\setlength{\belowcaptionskip}{-5pt}
\setlength{\intextsep}{5pt}

\section{Module Networks}

In the module network approach, the model learns a set of small networks (i.e., modules) and assembles them based on the question \cite{nmn}. The modules are described by a DSL, which specifies the arity and input/output types of the modules that the network has.

\subsection{NMN Architecture}

\begin{figure}[h!]
	\centering
	\includegraphics[width=0.8\linewidth]{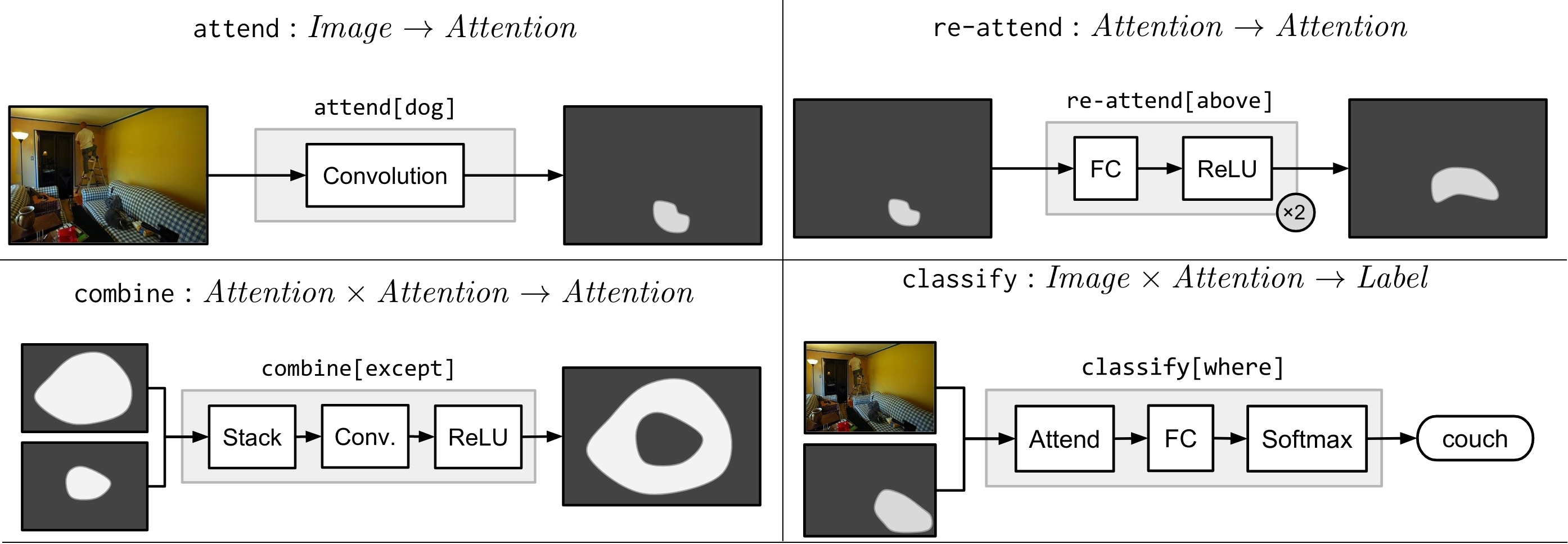}
	\caption{Graphic describing four of the five module types used by the NMN. Not shown is the \texttt{measure} module, which, given an attention map, returns a measurement (e.g., \texttt{measure[exists]}, or \texttt{measure[count]}). Modified from Andreas et al. \cite{nmn}}
	\label{fig:nmn}
\end{figure}

\begin{figure} [h]
	\centering
	\includegraphics[width=0.65\linewidth]{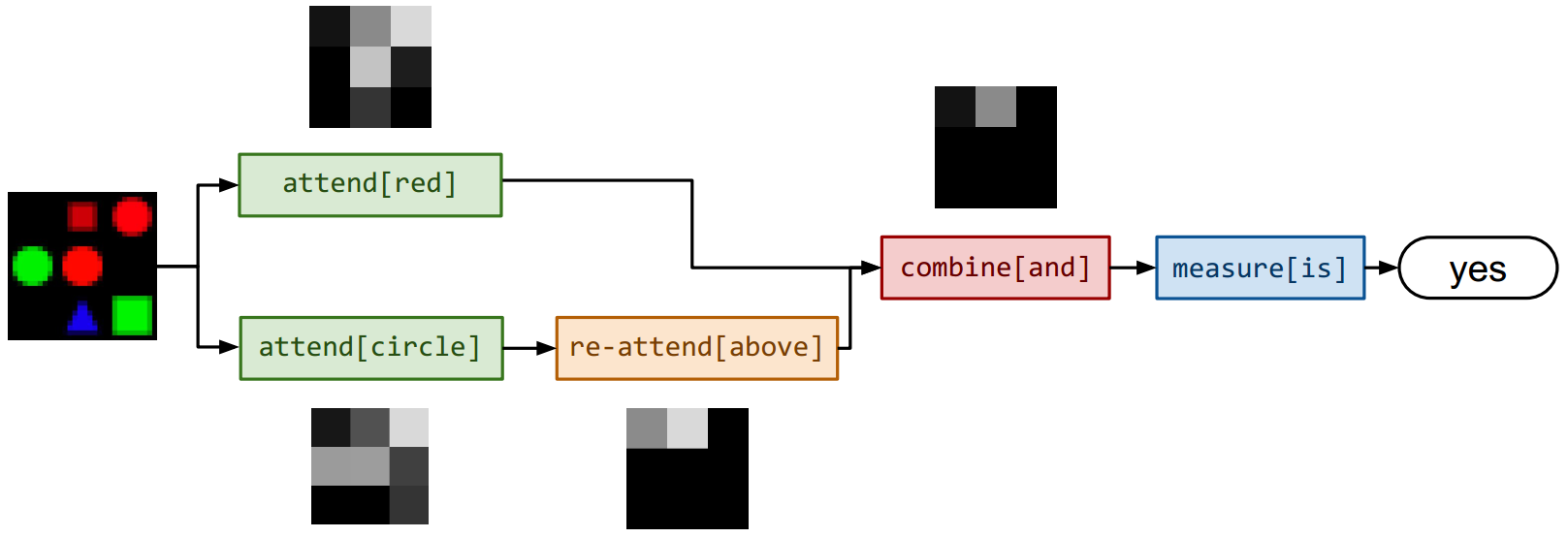}
	\caption{Visualization of NMN execution on a SHAPES problem. The question is ``Is there a red shape above a circle?'', and the corresponding program is \texttt{measure[is](combine[and](attend[red], re-attend[above](attend[circle])))}. Figure from Andreas et al. \cite{nmn}}
	\label{fig:nmnexec}
\end{figure}

The Neural Module Network (NMN) \cite{nmn} was the first module network for VQA. The NMN DSL lists 5 modules types (e.g., \texttt{attend}), each with distinct architectures (see Fig \ref{fig:nmn}). Most modules create or modify heatmaps of the image. Each module type has a unique instance per specific task. The sample execution in Fig \ref{fig:nmnexec} shows two such instances: \texttt{attend[circle]} and \texttt{attend[red]}. To produce the program, the NMN lemmatizes the question, applies the Stanford dependency parser \cite{9781905593507}, and applies fixed rules.

To overcome the limitations of the program generator and to incorporate a linguistic prior into the responses, the architecture above is ensembled with a blind LSTM model (i.e., predicts the answer based on the question alone). Specifically, the weighted geometric mean of the distributions is calculated; weights are determined by the text and image features. The system is trained end-to-end \cite{nmn}.

\subsection{NMN: Strengths \& Weaknesses}

The \textit{natural growth} of this architecture is critically limited by the immutable rules-based parser.  The authors note that the parser has problems; it aggressively simplifies the question (e.g., replacing plural nouns with singular nouns), and it introduces spurious predicates in roughly 10-20\% of VQA 1.0 questions (e.g., parsing \say{are these people most likely experiencing a work day?} as \texttt{be(people, likely)}). However, they perform no tests with ground-truth programs, even though SHAPES has ground-truth programs. 

In principle, the \textit{traceability} is good as the program is a trace, but there are caveats (see Section \ref{sec:fidelity}). \textit{Transfer learning} to another task by re-using modules is promising but untested. As each concept has its own independent module weights, the model cannot internally transfer acquired knowledge (e.g., learning \texttt{find[wolf]} using \texttt{find[dog]}), or exploit prior linguistic knowledge (e.g., via GloVe \cite{glove} or BERT \cite{bert} embeddings).

\begin{figure}[h!]
	\centering
	\includegraphics[width=0.2\linewidth]{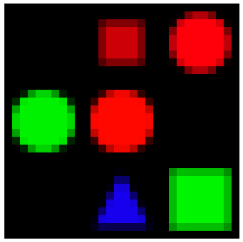}
	\caption{A sample image from the SHAPES dataset \cite{nmn}; the corresponding question is ``is there a red shape above a circle?''}
	\label{fig:shapes}
\end{figure}

The NMN exhibits \textit{compositional generalization} through new combinations of the learned modules. Andreas et al. \cite{nmn} tested the model's ability with the SHAPES dataset. However, although SHAPES tests complex compositional questions, all shapes are of the same size and aligned on a $3\times 3$ grid (see Fig \ref{fig:shapes}). This compensates for one of NMN's flaws, that reasoning steps, such as \say{look at the shape to the left}, are executed by a fixed \texttt{re-attend[left]} module without access to the input question or image, only the previous modules' outputs. Using a fixed grid simplifies this, e.g., \texttt{attend[above]} need not concern itself with objects that are above and of a different size, or occluded. 
The authors demonstrate that the model can generalize to SHAPES questions who's programs are one more module longer than seen at train time, but do not try combinations of subtasks that are new at test time (we discuss further in Section \ref{sec:closure}).

The NMN \cite{nmn} has no \textit{limitation awareness}; in fact, a common failure mode was to return a plausible response that is disconnected from the image. 
Although the NMN has no explicit confidence in its response, it indicates relative confidence between the module network and LSTM via the weights used in the ensemble. The model achieved SOTA performance on VQA 1.0 at the time of publication ($55.1\%$ accuracy). Although we now know that the dataset suffers from strong language priors \cite{balanced_binary_vqa, balanced_vqa_v2}, these priors were addressed through a blind baseline. 

\subsection{N2NMN: Improving the NMN}

\begin{figure}[h]
	\centering
	\includegraphics[width=1\linewidth]{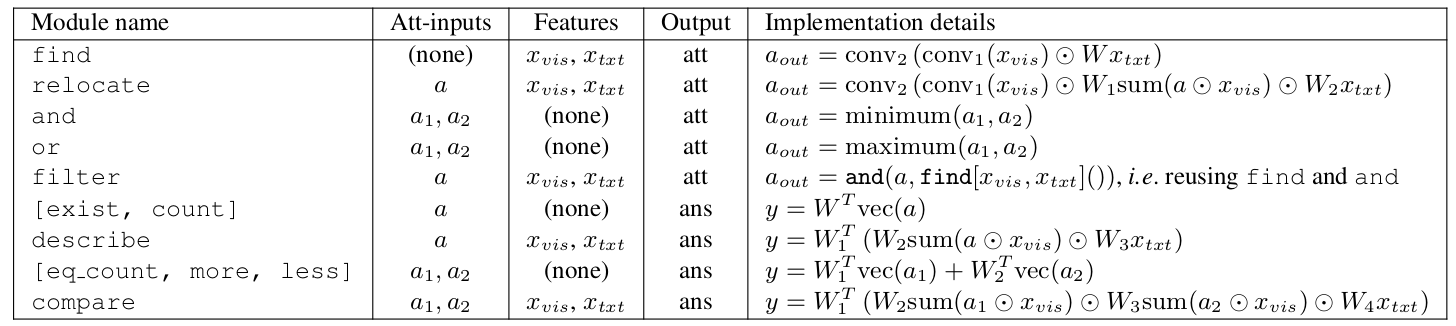}
	\caption{Table 1 of Hu et al. \cite{n2nmn} describing the N2NMN's modules, and their implementations. 
	}
	\label{fig:n2nmn}
\end{figure}

\begin{figure}[h!]
	\centering
	\includegraphics[width=1\linewidth]{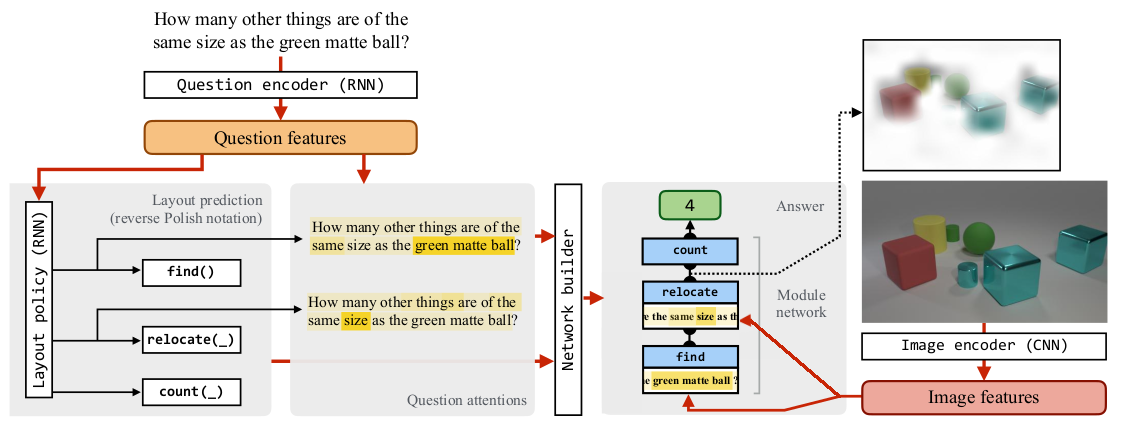}
	\caption{Illustration of N2NMN execution. Note that unlike the NMN which has separate instantiations for \texttt{attend[red]} and \texttt{attend[circle]}, the equivalent module in N2NMN, \texttt{find}, has only one instantiation. \texttt{find} chooses what to search for based on an $x_{txt}$ vector passed into the module; in this case $x_{txt}$ is a weighted average of the input tokens with most of the weight on ``green matte ball''. Figure from Hu et al. \cite{n2nmn}}
	\label{fig:n2nmn_exec}
\end{figure}

The End-to-End Module Network (N2NMN) \cite{n2nmn} followed the NMN. Primarily, it removed the blind model and produced programs (in reverse Polish notation) with a seq2seq encoder-decoder LSTM architecture. Additionally, each module type (e.g., \texttt{attend}) has exactly one instance and is provided a text-context vector, $x_{txt}$, (an attention over input tokens), and image features, $x_{vis}$, (extracted by VGG-16 \cite{vgg16}). The modules and their architectures are in Fig \ref{fig:n2nmn}, and Fig \ref{fig:n2nmn_exec} illustrates model execution.

Evaluation of the N2NMN on VQA 1.0 and SHAPES demonstrates increased performance over NMN. However, end-to-end training of the program generator through the modules via backpropagation is impossible. Executing the N2NMN requires arranging the modules; thus we must sample a discrete program from the program parser, and we cannot backpropagate through the sampling operation. REINFORCE can be used but causes a drop in performance ($4\%$ on SHAPES and $14\%$ on CLEVR). Instead, the N2NMN is trained in two stages: initially the generator is encouraged to mimic a provided expert parser, afterwards it is fine-tuned with REINFORCE. Thus training requires an expert parser to mimic.

\subsection{N2NMN: Strengths \& Weaknesses}

As the N2NMN's \cite{n2nmn} parser is trained on examples instead of being rule-based, the N2NMN improves on the \textit{natural growth} of the NMN. However, the N2NMN modules are still low-capacity, with architectures designed for the modules' intended role -- some modules such as \texttt{and} and \texttt{or} have no learnable parameters (see Fig \ref{fig:n2nmn}). Unless we violate compositionality by using multiple modules to implement a single operation, then each operation's capacity is severely limited.

The $x_{txt}$ vector passed to each module improves within-task \textit{transfer learning}, as each operation has a single module. Thus, for example, the 
\texttt{find} module can learn to find wolves based on its knowledge of dogs. However,  
$x_{txt}$ complicates \textit{transfer learning} to a different task; we require a mechanism to provide $x_{txt}$. $x_{txt}$ also reduces \textit{traceability}, as the program is coarser, and the user must interpret the attention defining $x_{txt}$ to guess what each module does. Furthermore, Subramanian et al. \cite{faithfulinterp} suggest that using $x_{txt}$ may damage \textit{compositional generalization} and \textit{traceability}, as it can leak information, causing hybrid operations. For instance, \say{red objects left of the cube} may correspond to \texttt{filter[red](relate[left](find[cube]))}. However, through $x_{txt}$, \texttt{relate[left]} may learn to find objects that are to the left \textit{and red}. This damages \textit{compositionality} (as \texttt{relate} will not behave as it should), and traceability (as the name \texttt{relate} becomes misleading if \texttt{relate} also does filtering); possible solutions, such as supervising intermediary modules outputs, are discussed in Section \ref{sec:fidelity}.

The \textit{compositionality} is tested by evaluation on CLEVR and SHAPES, however the authors do not test scenarios where the question distribution varies between train and test time. Other minor methodological flaws are that Hu et al. \cite{n2nmn} omitted the performance when training without program supervision on VQA 1.0, and they augmented CLEVR image features with two channels for position data, without an ablation test. Like the NMN, the N2NMN has no \textit{limitation awareness}.

\subsection{Compositional Generalization and Few-Shot Learning in Module Networks}\label{sec:closure}

Underpinning module networks is the idea that we can solve new combinations of concepts with new combinations of modules. We can test this assertion through CLOSURE \cite{closure}, a dataset for testing the \textit{compositional generalization} of models trained on CLEVR. Where CLEVR's test questions are generated from one of 90 templates seen at train time 
\cite{clevr}, CLOSURE is generated from five new templates created by replacing spatial comparisons with attribute comparisons. 

The authors used CLOSURE to test the \textit{compositional generalization} of various SOTA models, and the effectiveness of \textit{few-shot learning}. Critically, the NMN-type model tested (dubbed Tensor-NMN \cite{johnson2017inferring}) failed to generalize, even with ground-truth programs. The Tensor-NMN architecture's modules take and return tensors of fixed dimensions; all modules are residual blocks of two $3\times 3$ convolutions, with higher arity modules first merging input channels with $1\times1 $ convolution. A program parser similar to the N2NMN's is used. The Tensor-NMN DSL is similar to NMN's, with multiple types that have different instantiations (e.g., \texttt{filter\_colour[blue]}, \texttt{filter\_colour[red]}, etc...). This demonstrates that the N2NMN parser is problematic, and that the N2NMN and NMN modules may not function as intended when rearranged.

The authors speculated that reducing the dimensionality of the intermediary features would improve generalization. They thus developed Vector-NMN, which only passes vectors between modules. Vector-NMN uses the same program parser as Tensor-NMN and a different module architecture. All modules are  two consecutive FiLM \cite{film} blocks applied to ResNet101 features of the input image, followed by max-pooling to reduce them to a vector. The FiLM blocks contain convolutional filters shared among all modules, which are modulated with additive and multiplicative weights produced by a MLP. This MLP processes the vector outputs of any previous modules along with a learned module-embedding representing the module's role. Bahdanau et al.\cite{closure} found that Vector-NMN had near perfect generalization to CLOSURE when given the ground-truth programs, with the exception of one family of questions. Here, Vector-NMN still outperformed Tensor-NMN. However, we cannot conclude that dimensionality alone determines generalization; the authors overlook key differences between the models. Notably: Vector-NMN uses weight sharing, is architecturally more similar to FiLM than Tensor-NMN, and its modules get image features as an argument. Comparison with other models using lower dimensional intermediary values such as TbD \cite{TbD} (mostly matrices) and Visual-NMN \cite{faithfulinterp} (vectors) would help with the confounding variables.

Although 
self-evident, the largest limitation of CLOSURE is that it is a necessary, but insufficient, condition for \textit{compositional generalization} on CLEVR. For instance, all CLOSURE questions contain the same number of 
referring expressions as some CLEVR question. Consequently, CLOSURE only tests compositional generalization on questions that are at most as complex as CLEVR questions. 

The authors also test \textit{few-shot learning} by training Vector-NMN on 36 questions from each CLOSURE template without program annotations. The findings were that in six of the seven families the model performed well after training, demonstrating that the seq2seq LSTM program-parser can exhibit \textit{few-shot learning} using REINFORCE only. However, fine-tuning failed on the seventh family; the authors concluded that it was because REINFORCE was not sampling good programs. This suggests that there may be scalability issues 
as more complex DSLs are used.

\subsection{Fidelity in Module Networks}\label{sec:fidelity}

\begin{figure}[h!]
	\centering
	\includegraphics[width=0.7\linewidth]{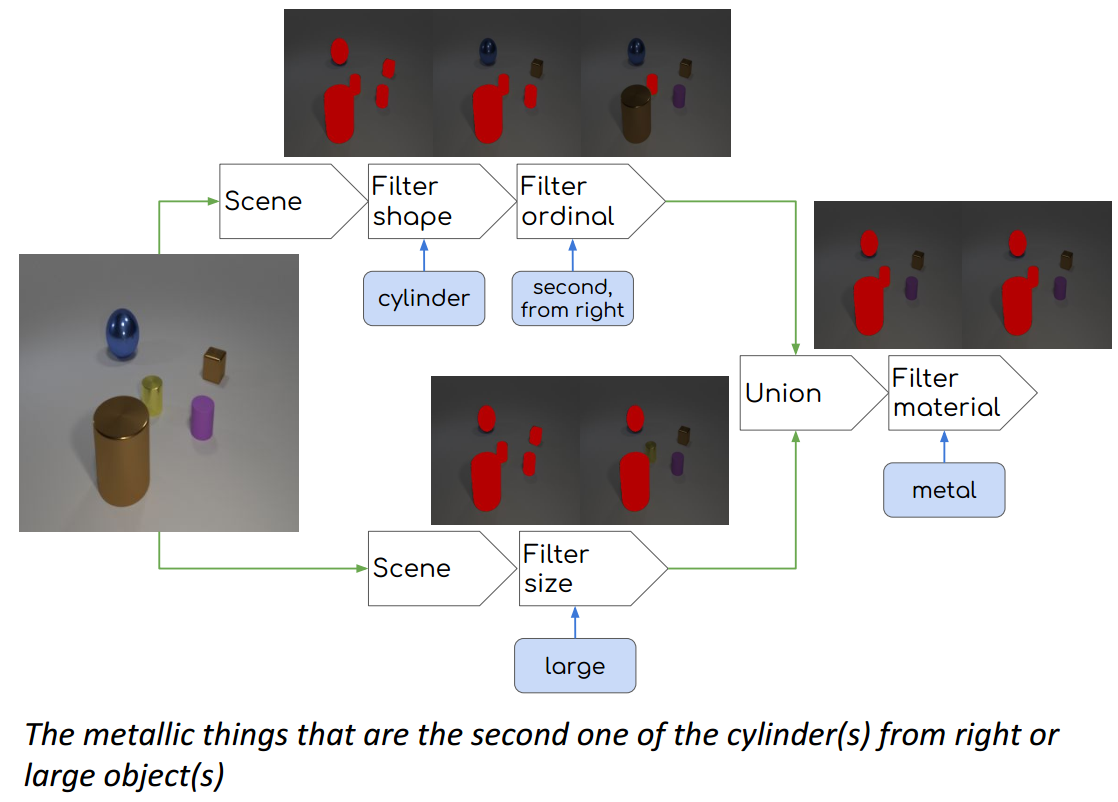}
	\caption{Figure illustrating execution of Liu et al.'s \cite{CLEVR-Ref+} Tensor-NMN modified for image segmentation. Note that not shown is a final module, \texttt{postprocess}, that creates the final segmentation. Note also, that the segmentations shown are \textbf{not} the features passed from module to module; Tensor-NMN's pass tensors between modules. The shown images are the results of passing these intermediary tensors into the \texttt{postprocess} module. Note the high fidelity to what we would expect to see out of the partial operations. Note also that the blue boxes are \textbf{not} arguments. Like the NMN, the Tensor-NMN and it's variants have a separate instantiation of each module for each possible value in the blue boxes. Figure from Liu et al. \cite{CLEVR-Ref+}}
	\label{fig:clevr-ref}
\end{figure}

Failure to generalize with ground truth programs calls into question the faithfulness of modules' behaviour to the program DSL, and thus the \textit{traceability}. Liu et al. \cite{CLEVR-Ref+} found that Tensor-NMN modified for image segmentation exhibited good fidelity, and that the shared tensor output dimensions of all modules facilitated \textit{transfer-learning} (see Fig \ref{fig:clevr-ref}). Specifically, they could apply the last module, which produces the output segmentation heatmap, to any intermediate module, thus producing an intermediate image segmentation. The modules' fidelity was not perfect; 
one module ultimately functioned as a preprocessing step instead of the desired behaviour. Similarly, Subramanian et al. \cite{faithfulinterp} found poor fidelity in a module network using probabilistic sets as intermediary values. They found that restricting module architectures or supervising module outputs improved fidelity; this creates 
a tradeoff between improving \textit{traceability} through fidelity, and reducing \textit{natural growth} due to lower capacity modules, or training requiring additional annotations.

\section{Addressing the Problems of Discrete Programs}
\subsection{Reducing model capacity}

\begin{figure}[h!]
	\centering
	\includegraphics[width=1\linewidth]{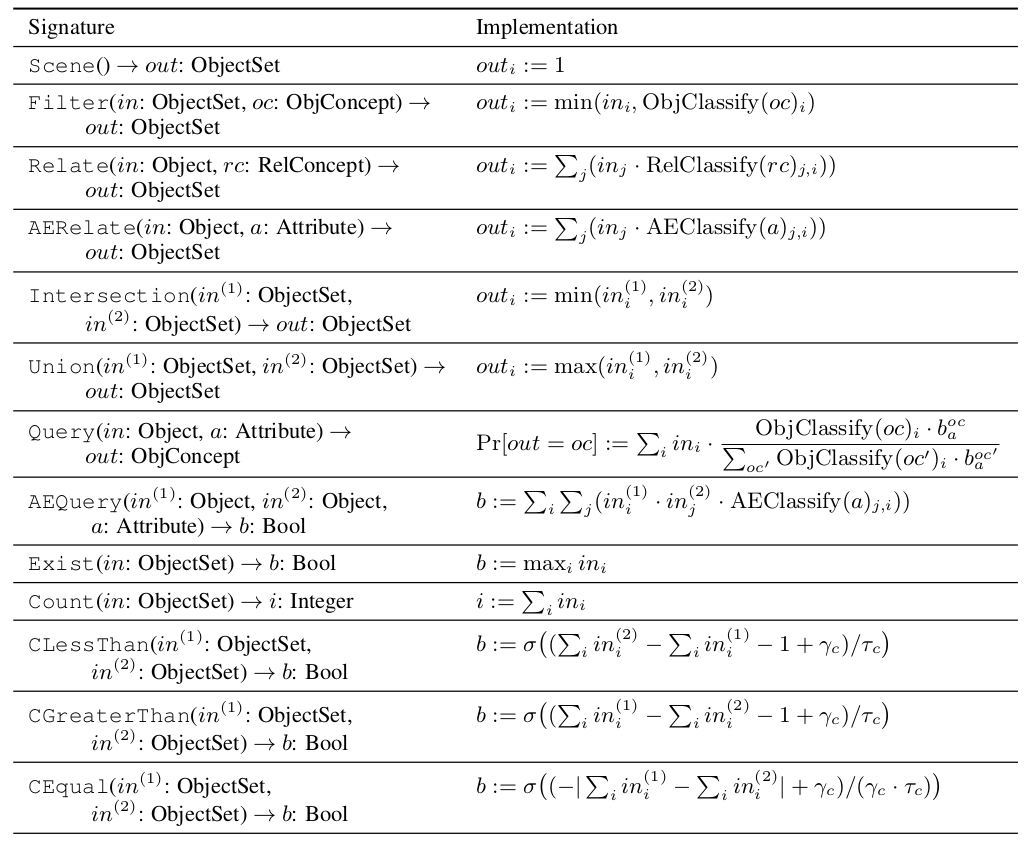}
	\caption{Table describing the NS-CL's DSL and implementation. Note that $\tau_c = 0.25$ and $\gamma_c = 0.5$ are hyperparameters. Note that given the set of all attributes, $A$, then for every concept $oc$ (e.g., red), we must be given a 1-hot vector $b^{oc} \in \mathbb{R}^{|A|}$ indicating the attribute $a \in A$ that this concept belongs to. With this given constant, then for objects $i$ and $j$, and attribute $a$ (e.g., colour) with concept $oc$ then the probability that object $i$ is $oc$ is estimated by $\text{ObjClassify}(oc)_i = \sigma \left( \sum_{a \in A} \left( b_a^{oc} \frac{\langle u^a(o_i), v_{oc} \rangle - \gamma}{\tau} \right) \right)$ where $u^a$ is the embedding MLP for the attribute $a$, $o_i$ is the feature vector for object $i$, and $v_{oc}$ is the learned vector representation of the concept $oc$. Similarly, the probability that $i$ and $j$ have the same attribute $a$ is estimated by $\text{AEClassify}(a)_{ij} = \sigma\left( \frac{\langle u^{a}(o_i), u^a(o_j) \rangle - \gamma}{\tau} \right)$. The probability that objects $i$ and $j$ are related by the binary relationship $rc$ is estimated by $RelClassify(rc)_{ij} = \sigma \left( \frac{\langle u^r(o_i, o_j), v_{rc} \rangle - \gamma}{\tau} \right)$ where $u^r$ is the embedding MLP for all binary relationships, and $v_{rc}$ is the learned vector representation of the binary relation $rc$. Note $\gamma$ and $\tau$ are again hyperparameters. Table from Mao et al. \cite{ns-cl}}
	\label{fig:nscl}
\end{figure}

The Neuro-Symbolic Concept Learner (NS-CL) resolves the module composition and fidelity issues by using fixed implementations for most elements of the DSL (see Fig \ref{fig:nscl}). This ensures perfect \textit{traceability}, and some \textit{limitation awareness} to the model; ambiguous or invalid questions cause the execution engine to throw an error. However, this comes at the cost \textit{natural growth}; the DSL's execution is fixed.

\begin{figure}[h!]
	\centering
	\includegraphics[width=1\linewidth]{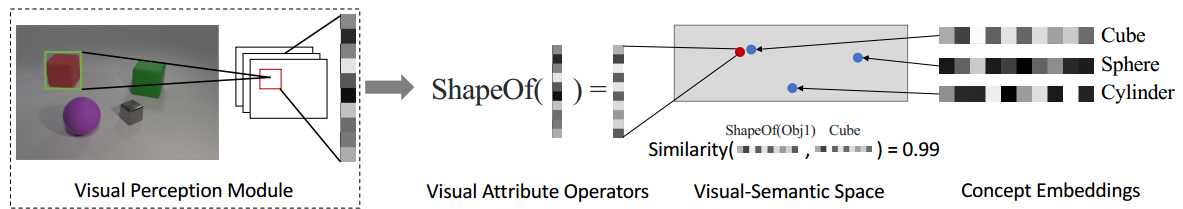}
	\caption{Illustration of the learned components of the NS-CL's exector. The only components of the executor that are learned are concept embeddings (e.g., the cube, sphere, and cylinder embeddings for the shape attribute), and MLP's that project object representations from the Mask R-CNN into the same attribute space as the corresponding concepts (in this case, projecting object representations into shape space). Figure from Mao et al. \cite{ns-cl}}
	\label{fig:nscl-learned}
\end{figure}

Instead of processing attention maps, the NS-CL \cite{ns-cl} identifies objects with a pre-trained Mask R-CNN \cite{mask_r_cnn} and processes probabilistic sets of objects. Each object has a vector representation of ResNet-34 features. The execution engine's learned components are: i) networks for projecting an object's feature vector into an attribute space, and ii) a vector in said attribute space for each concept specified by the user (see Fig \ref{fig:nscl-learned}). This structured representation allows for some \textit{transfer learning} to other tasks. The authors demonstrated this by freezing the attribute embedding networks and using them with a different DSL for CLEVR image retrieval.

The parser architecture is also changed, and a rule-based preprocessing step is added. Concepts \& attributes in the question are replaced with placeholders, thus allowing new attributes to be recognized by updating the list of attributes. This combined with the disentangled attribute representations facilitates \textit{few-shot learning}; 
Mao et al. \cite{ns-cl} show that when learning a new colour from $100$ example images, the NS-CL has a test accuracy of 93.9\%, outperforming TbD \cite{TbD} and Tensor-NMN by $6.1 \%$ and $4.6 \%$ respectively.  Furthermore, if attribute detection could be automated, then this would facilitate \textit{natural growth} to learn new attributes and concepts. The authors do not report a seq2seq baseline.

Like module networks, the design promises \textit{compositional generalization}. The authors test on CLEVR and CLEVR-CoGenT, demonstrating the NS-CL is capable of answering compositional questions and exhibits compositional generalization under a change of attribute combinations. 
The authors claim that NS-CL can generalize to new questions, and support this by training and testing on a split of CLEVR based on program depth. This implicitly splits by CLEVR question template, and is thus similar to CLOSURE.

On a practical note, the additional structure improves performance on CLEVR compared to the SOTA module network approaches such as TbD \cite{TbD}. Furthermore, the NS-CL does not require program annotations at train time and is more data-efficient than TbD. Curriculum learning is used to train the model, 
and an ablation demonstrated that removing curriculum learning resulted in either non-convergent training, or performance on-par with random guessing \cite{ns-cl}. 
The first stage 
of training 
consists of short questions about attributes, the second stage adds short relational questions, and the third stage 
adds all remaining questions. 
During the first two stages, REINFORCE is used to train the program parser. In stage three, the exact REINFORCE gradient is calculated by finding $\mathbb{Q}$, the set of programs returning the correct response. The attribute embedding networks and concept vectors are trained simultaneously through backpropagation, although they are initially frozen in stage 3. However, finding the true $\mathbb{Q}$ is intractable. After writing to the authors, I determined that they restricted their search to programs from templates extracted from ground-truth programs. This indicates that \textit{scalability} with DSL complexity is problematic, and the severity is unclear 
as the authors do not report performance when using normal REINFORCE in stage 3.

\subsection{Continuous Programs}

\begin{figure}[h!]
	\centering
	\includegraphics[width=1\linewidth]{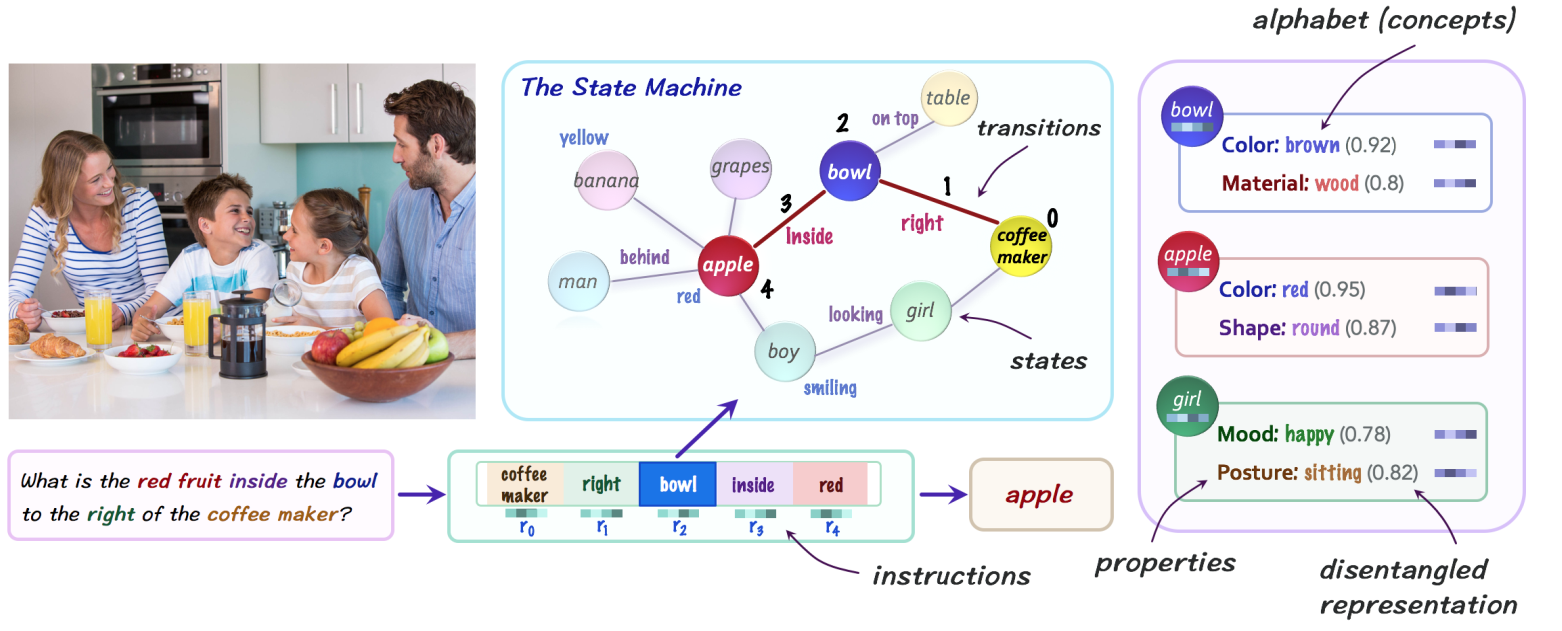}
	\caption{Illustration of NSM execution. First, the question is converted into a fixed number of continous reasoning instructions (in this case $r_0, r_1, \dots r_4$), and the image is converted into a scene graph. Next, a uniform attention is placed over the nodes, and sequentially updated using the reasoning instructions. Updates are a linear combination of an updated based on node similarity (e.g., finding the coffee machine when executing $r_0$), and edge similarity (e.g., following the edge from the coffee machine to the bowl when executing $r_1$). Figure from Hudson \& Manning \cite{nsm}}
	\label{fig:nsm-exec}
\end{figure}

Where the NS-CL addresses the issues arising from discrete programs with a (mostly) fixed execution engine, the Neural State Machine (NSM) instead resolves these issues by using continuous programs. Although it doesn't use a DSL, the NSM still incorporates prior knowledge in the form of the object-types, concepts, attributes, and binary relationships that can occur in an image. 
Specifically, NSM performs soft reasoning over a graph representation of the image (see Fig \ref{fig:nsm-exec}). 
This design hurts \textit{traceability}, as the reasoning 
steps are now vectors. 
At best, the meaning of these reasoning instructions can be guessed based on the attention over the question tokens that created them.

To create this graph representation, the NSM requires: the set of object-types, the attributes, a set of concepts for each attribute, and the set of binary relationships. First, a node is created in the graph for each object detected by a Mask R-CNN. This Mask R-CNN also produces distributions over object type and attributes; this requires scene graphs at train time. Next, for all objects that are sufficiently close in the image a directed edge is created in each direction between the corresponding nodes. A graph attention network \cite{gat} then produces a distribution over binary relationships for each edge. Each distribution is summarized as a weighted average over learned concept/object/relationship embeddings, initialized with GloVe. This graph is the final representation of the image that the NSM reasons over. The NSM redistributes attention over the nodes based on the reasoning instruction, sometimes through edges.

This design severely limits the \textit{natural growth} of the model compared to the approaches seen thus far. Previously, another attribute or concept could be added by the addition of a module (e.g., NMN), or the updating of a list (e.g., NS-CL); for the NSM this requires changing the model's architecture to alter the arity of various networks. However, the discrete scene graph representation does allow for the NSM to be cleanly divided into two components that can be reused in other tasks, potentially allowing for \textit{transfer learning}, albeit in a manner that is coarser than module reuse in a module network. The fixed edge creation heuristic is also problematic as the model cannot reason about long-distance relationships (e.g., between a thrower and catcher on a game field).

The flow of attention through the graph is controlled by \textit{reasoning instructions} that are generated from the question, somewhat analogous to 
the N2NMN's $x_{txt}$ vectors. 
The reasoning instructions are weighted averages of vectors representing the input tokens of the question, and are produced by an encoder-decoder model. However, where the N2NMN learns token embeddings, the NSM represents each token as a weighted average between the  
GloVe embedding and the learned object, concept, and relationship embeddings.
Ablation indicates that representing both the question and image as weighted averages of the same embeddings improves performance. Furthermore, the NSM always decodes a fixed number of reasoning instructions, thus allowing the parser to be trained through backpropagation. 

Starting from uniform attention over nodes, the NSM sequentially updates the attention based on each reasoning instruction. The update is a linear combination of two mechanisms; one assigns attention 
based on attributes and object-type, the other shifts attention 
along edges based on binary relation. 
Final classification is done by a 2-layer MLP given the encoded question (from the encoder-decoder reasoning instruction parser), and a vector representation of the graph. The graph representation is produced by first averaging each node's attribute vectors based on the last reasoning instruction, and then averaging these vectors based on the final attention over nodes. 

The authors demonstrate the NSM's \textit{compositional generalization} through 
splits of GQA where object types are removed at train time and introduced at test time. Note that object detector is trained on all objects. However, GQA questions are generated from 
templates, and the authors do not have a CLOSURE-like test for generalization to new question structures.

\section{Achieving Desiderata via Regularization}

The previous models incorporated prior symbolic knowledge directly into their reasoning processes; an alternative approach is to instead leverage this knowledge in model optimization. Xie et al. \cite{lensr} defines a regularizer, $\ell_{logic}$, 
that encodes propositional logic describing the domain (e.g., \texttt{wear(person,hat)} $\implies$ \texttt{exists(person)} $\wedge$ \texttt{exists(hat)}). 
Note that unlike 
the previously discussed models, the authors tested on the related task of visual relationship prediction (VRP): given an image, the bounding boxes of two objects, and their respective object-type, VRP seeks the relationship between the objects.

The regularizer uses a trained Logic Embedding Network with Semantic Regularization (LENSR). 
To train a VRP model, $h$, the first step is to train LENSR, $q$, on the given rules, and then freeze the LENSR weights. LENSR is a modified graph convolutional network \cite{gcn} that encodes propositional logic formulas as a vector. To encourage the encoding of assignments to be closer to formulas that they satisfy, LENSR is trained with a regularized triplet loss (details in Appendix \ref{sec:lensrDetails}). $h$ is then trained with cross entropy loss regularized with $\twonorm{q(F) - q(\bigwedge_i p_i)}^2$. $F$ is the formula of rules (in d-DNNF form) that apply to this batch of training examples, and $p_i$ is the weighted average of the LENSR embeddings of all possible solutions for the $i$th example in the batch, weighted by $h$ (see Fig \ref{fig:lensr}).

LENSR demonstrates an approach for \textit{transfer learning} by which pre-existing symbolic knowledge guides training. Systems that automatically extract symbolic knowledge from text, such as COMET \cite{comet}, 
may thus allow \textit{natural growth} from unstructured data. However, there are limitations. LENSR encodes propositions by summing the GloVe embeddings of their components (e.g., \texttt{on(sphere, block)} $\rightarrow$ \texttt{block}+\texttt{on}+\texttt{sphere}), thus loosing order information. 
Furthermore, LENSR may not scale well as 
compiling the rules into d-DNNF form is an NP-Hard task.

As to practical considerations, the authors demonstrated that the regularization increases model performance on VRP when $h$ is a two-layer MLP, and that their logic loss outperforms semantic loss \cite{semanticLoss}; they speculate this due to the more complex constraints. They also compare with a TreeLSTM encoder, however previous work using TreeLSTM for logical entailment was under different conditions (see Appendix \ref{sec:lensrmethod}). 

\section{Discussion and Conclusions}

In this work, we discuss how various neuro-symbolic VQA models meet a set of AGI desiderata. We find that, in principle, module networks satisfy all the desiderata except for \textit{limitation awareness}. However, \textit{transfer} to significantly different tasks is untested, and \textit{traceability} is closely tied to program fidelity, as is \textit{compositionality}. Yet, approaches that encourage program fidelity, such as 
parameter-free modules, often harm \textit{natural growth} -- with the extreme example being the NS-CL. Furthermore, CLOSURE indicates that encoder-decoder program parsers do not exhibit \textit{compositional generalization} to new combinations of sub-tasks. The NSM avoids the program parser generalization and program supervision problems and achieves the best GQA performance, but at the cost concessions on \textit{traceability}, \textit{transfer learning} and \textit{natural growth}. 
LENSR suggests that the choice of training procedure may be able to encourage AGI desiderata.  

If we seek to overcome the limitations of module networks, then we may be able to improve parser generalization through the training procedure. In addition to NS-CL style curriculum learning, we may be able to apply techniques from domain-generalization by framing families of questions from specific templates as domains. A significant open question about a module network's \textit{natural growth} is whether the DSL can be learned from data; it's possible that a more structured intermediary representation like the NSM may simplify this task -- although it may be a lateral move, placing the onus of growth on the scene graph.

Overall, the existence of models exhibiting subsets of the desiderata, combined with the nice interpretation of many of the desiderata in VQA (\textit{compositionality} as new combinations of old ideas, \textit{natural growth} as adaption to new questions and domains, \textit{traceability} as the production of accurate traces, \textit{transfer learning} within-task as exploiting old knowledge to learn new concepts, and out of task in the usual sense, and \textit{self awareness of limitations} as identifying invalid questions, or indicating confidence), indicates that VQA may be a rich task for the purposes of developing models that exhibit all of these desiderata. Such work will not bring us AGI, but can provide a clearer image of the challenges that lie on the road there.

\clearpage

\section{Appendix}

\setlength{\abovecaptionskip}{5pt}
\setlength{\belowcaptionskip}{5pt}
\setlength{\intextsep}{20pt}

\subsection{Sample Questions from VQA Datasets}

\clearpage

\subsection{DSLs}

\clearpage

\subsection{Sample Execution}

\begin{figure}[h!]
	\centering
	\includegraphics[width=1\linewidth]{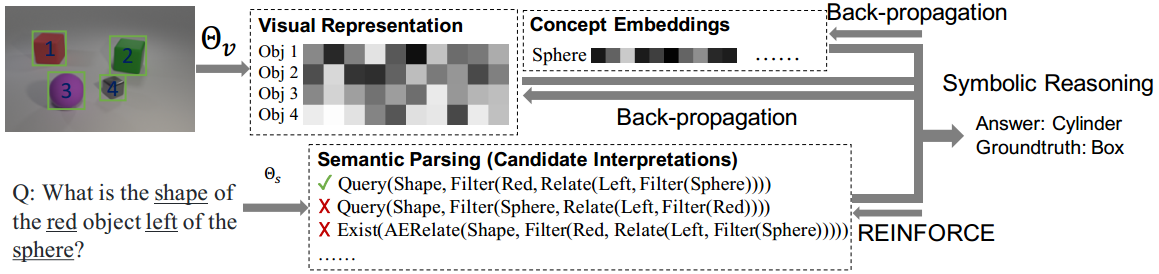}
	\caption{Illustration of NS-CL execution and training. First the Mask R-CNN identifies the objects in the image and produces object representations. Independently the program parser converts the natural language question into a program. This program and the object representations are then passed into the deterministic program executor. The only learned components of the executor are the concept embeddings, and attribute embedding networks that project object representations into the same attribute space as the concept embeddings corresponding to the attribute. Figure from Mao et al. \cite{ns-cl}}
	\label{fig:ns-cl-exec}
\end{figure}

\begin{figure}[h!]
	\centering
	\includegraphics[width=1\linewidth]{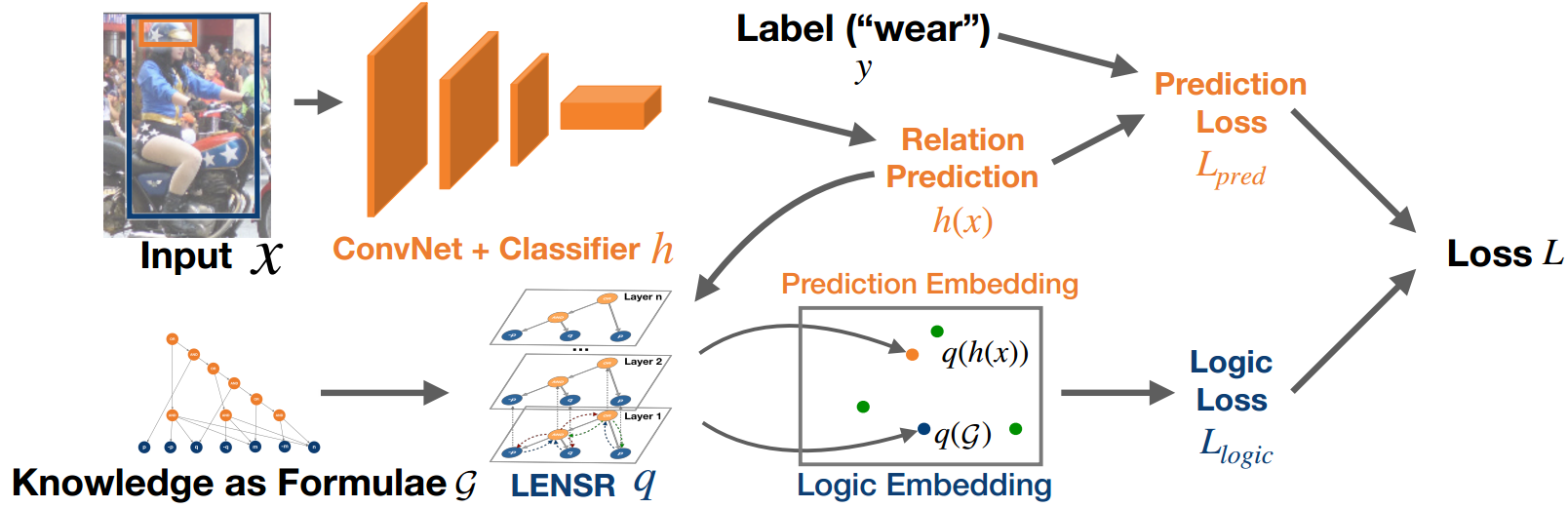}
	\caption{Illustration of the training of a VRP model $h$ using the logic loss regularizer, $L_{logic}$, defined w.r.t. LENSR, $q$. The logic loss regularizer is the squared Euclidean distance between $q(\mathcal{G})$ (the LENSR encoding of the logical formula describing the domain), and $q(h(x))$ (the average over the LENSR encodings of all possible solutions, weighted by the model undergoing training, $h$). At this point LENSR is frozen, and has already been trained to reduce the distance between the encodings of formulas and their satisfying assignments. Figure from Xie et al. \cite{lensr}}
	\label{fig:lensr}
\end{figure}

\clearpage

\subsection{LENSR Details}\label{sec:lensrDetails}

The LENSR logic loss is dependent on encoding prepositional logic expressions as vectors \cite{lensr}. The first step is to convert the expression into d-DNNF form. This form is attractive as it allows for clausal entailment, model counting, model enumeration, model minimization based on cardinality, and probabilistic equivalence testing to be solvable in time polynomial with the size of the d-DNNF formula \cite{ecai2004}. Note that as a consequence, conversion from CNF to d-DNNF is NP-Hard \cite{ecai2004}. d-DNNF form is both deterministic (the arguments of any disjunction, $\vee$, are mutually inconsistent) and decomposable (the arguments of any conjunction, $\wedge$, are over disjoint sets of variables) \cite{ecai2004}. 

Once the formula is in d-DNNF form, it is encoded into a vector by LENSR. LENSR is a modified graph convolutional network (GCN) \cite{gcn}. Before processing by LENSR, the d-DNNF formula is converted into a non-binary expression tree (see $\mathcal{G}$ in Fig \ref{fig:lensr}), the edges are made undirected, and a ``global'' node is created that connects to all other nodes. This graph is then processed by the LENSR, and the final LENSR embedding of the global node is used as the final embedding of the formula.

Where a standard GCN shares the same weight matrix across all nodes, LENSR is modified to use \textit{heterogeneous node embedding}. In other words, there is a distinct weight matrix used for $\vee$ nodes, $\wedge$ nodes, leaf nodes, and the global node. To train LENSR, the loss function $\sum_F \sum_{\tau_f, \tau_t} \ell_t(F, \tau_f, \tau_t) + \lambda \ell_r(F)$ is used, where $\lambda$ is a hyperparameter, $F$ is a d-DNNF formula in the training data, $\tau_f$ is a non-satisfying assignment (expressed as the conjunction of literals for all variables in $F$) and $\tau_t$ is a satisfying assignment. For example, we could have $F = (A \wedge B) \vee (\neg A \wedge \neg B)$, $\tau_t = A \wedge B$, and $\tau_f = A \wedge \neg B$. The term $\ell_t$ is a triplet loss encouraging satisfying assignments to be closer to the formula than non-satisfying assignments. Specifically, $\ell_t = \max \{ \twonorm{q(F)-q(\tau_t)}^2 - \twonorm{q(F)-q(\tau_f)}^2 + m, 0\}$ where the margin, $m$, is a hyperparameter, and $q$ is the LENSR network. The other term ($\ell_r$) is the semantic regularization component, which encourages properties in the latent space analogous to properties of the d-DNNF. At a high-level, the embeddings of the children of an $\vee$ node should add to $\mathbf{1}$ (as an analogue to mutual exclusivity), and the embeddings of the children of an $\wedge$ node should be orthogonal (as an analogue to operating on disjoint sets of variables). Specifically, $\ell_r = \sum_{v_i \in \mathcal{N}_O}\twonorm{\mathbf{1} - \sum_{v_j \in \mathcal{C}_i} q(v_j) }^2 + \sum_{v_k \in \mathcal{N}_A}\twonorm{V_k^T V_k - \text{diag}(V_k^T V_k)}^2$, where $\mathcal{N}_O$ is the set of $\vee$ nodes, $\mathcal{N}_A$ is the set of $\wedge$ nodes, $\mathcal{C}_i$ is the set of children nodes of $v_i$ in the directed expression tree, and $V_k := [q(v_1), \dots q(v_{l_k})]$ s.t. $v_i \in \mathcal{C}_k$ and $l_k = |\mathcal{C}_k|$, .

Note that the graph convolutional network requires initial embeddings for all nodes. Non-leaf nodes are given an initial embedding based on their type ($\vee$, $\wedge$, or global), which is shared among all nodes of the same type. The initial embeddings for leaf-nodes are created by taking the sum of the GloVe embeddings of the words comprising the proposition. For example, the proposition \texttt{wear(person,glasses)} is represented as the sum of the GloVe embeddings for \say{wear}, \say{person}, and \say{glasses}.

\subsection{LENSR Methodological Flaw}\label{sec:lensrmethod}

Xie et al. \cite{lensr} test their model on the VRD dataset \cite{vrd}, and compare against TreeLSTM \cite{treelstm} encoding. They state that TreeLSTM \cite{treelstm} is state-of-the-art. However, to the best of my knowledge, TreeLSTM has never been used on VRD before. Furthermore, the paper they cite concerns encoding natural language rather than logic. Reviewing the references, I found a paper (not cited within the body) demonstrating that for the task of logical entailment, TreeLSTM was the best performing encoding-type model \cite{evans2018can}. From this, I assume that authors intended to convey that TreeLSTM was SOTA for determining entailment using embeddings. However, Evans et al. \cite{evans2018can} used a modified variant of TreeLSTM where parameters at each node are determined by the corresponding logical operators, and it is unclear whether Xie et al. \cite{lensr} used the modified or unmodified TreeLSTM. Furthermore, the cited paper used a MLP to determine entailment based on the encodings, as opposed to squared euclidean distance used with the LENSR embeddings. It is unclear how the LENSR authors used the TreeLSTM embeddings, which may make their comparison unfair.

\printbibliography

\end{document}